\documentclass[letterpaper]{mc2023}
\usepackage{tabls}
\usepackage{cites}
\usepackage{epsf}
\usepackage{appendix}
\usepackage{ragged2e}
\usepackage[top=1in, bottom=1in, left=1in, right=1in]{geometry}
\usepackage{enumitem}
\setlist[itemize]{leftmargin=*}
\usepackage{caption}
\usepackage{newtxmath}
\usepackage{booktabs}
\usepackage{amsmath}
\usepackage{bm}
\usepackage{multirow}
\usepackage{multicol}
\usepackage{graphicx}
\usepackage{subfig}
\usepackage[justification=centering]{caption}
\usepackage{multicol}
\usepackage{multirow}
\graphicspath{{./figs/}}

\title{Reduced Order Modeling of a MOOSE-based Advanced  \\Manufacturing
Model with Operator Learning}

\author{
  \textbf{Mahmoud Yaseen$^1$, Dewen Yushu$^2$, Peter German$^3$ and Xu Wu$^1$}\vspace{3pt} \\
  $^1$Department of Nuclear Engineering, North Carolina State University  \\
  Burlington Engineering Laboratories, 2500 Stinson Drive, Raleigh, NC, USA 27695 \vspace{3pt}\\
  $^2$Computational Mechanics \& Materials Department \\
  Idaho National Laboratory, P.O. Box 1625, Idaho Falls, ID 83415 \vspace{3pt}\\
  $^3$Computational Frameworks Department \\
  Idaho National Laboratory, P.O. Box 1625, Idaho Falls, ID 83415 \vspace{3pt}\\
  \url{mqyaseen@ncsu.edu}, \url{Dewen.Yushu@inl.gov}, \url{Peter.German@inl.gov}, \url{xwu27@ncsu.edu}
}

\newcommand{\authorHead}{Yaseen, Yushu, German and Wu}
\newcommand{\shortTitle}{ROM of a MOOSE-based AM Model with Operator Learning}

\begin{document}
\maketitle
\pagestyle{fancy} \cfoot{\thepage}
\fancypagestyle{firstpage}{\fancyhead[C]{\footnotesize{\emph{
M\&C 2023 - The International Conference on Mathematics and Computational Methods Applied \\
to Nuclear Science and Engineering $\cdot$ Niagara Falls, Ontario, Canada $\cdot$ August 13 -- 17, 2023}}}
\cfoot{}}
\thispagestyle{firstpage}
\fancyhead[CE]{{\scriptsize \authorHead}}
\fancyhead[CO]{{\scriptsize \shortTitle}}
\justify 
\parskip 6pt plus 1 pt minus 1 pt

\begin{abstract}
  Advanced Manufacturing (AM) has gained significant interest in the nuclear community for its potential application on nuclear materials. One challenge is to obtain desired material properties via controlling the manufacturing process during runtime. Intelligent AM based on deep reinforcement learning (DRL) relies on an automated process-level control mechanism to generate optimal design variables and adaptive system settings for improved end-product properties. A high-fidelity thermo-mechanical model for direct energy deposition has recently been developed within the MOOSE framework at the Idaho National Laboratory (INL). The goal of this work is to develop an accurate and fast-running reduced order model (ROM) for this MOOSE-based AM model that can be used in a DRL-based process control and optimization method. Operator learning (OL)-based methods will be employed due to their capability to learn a family of differential equations, in this work, produced by changing process variables in the Gaussian point heat source for the laser. We will develop OL-based ROM using Fourier neural operator, and perform a benchmark comparison of its performance with a conventional deep neural network-based ROM.
\end{abstract}

\vspace{6pt}

\section{INTRODUCTION}

Recently there has been a significant interest in advanced manufacturing (AM) across the US DOE and the nuclear industry. AM is a broad concept encompassing all technologies that produce parts via a layer-by-layer process, such as fused deposition modeling, stereolithography, direct energy deposition (DED), and powder bed fusion, etc. In this work, we will focus on DED, which has also been called laser-engineered net shaping (LENS), direct metal deposition, and laser consolidation. Recently, a novel, geometry-free thermo-mechanical model has been developed \cite{yushu2022directed} at the Idaho National Laboratory (INL), which uses adaptive subdomain construction to accurately predict the thermal conditions, distortions, and residual stresses throughout the DED process. This model is based on the open-source Multiphysics Object-Oriented Simulation Environment (MOOSE) framework \cite{lindsay20222} developed at INL. The goal of this work is to build an accurate and fast-running reduced order model (ROM) for this high-fidelity MOOSE-based AM model.

One major challenge for AM application in nuclear materials is to obtain desired properties via controlling the manufacturing process during runtime. The existing hardware level controllers (e.g., proportional integral derivative) are designed to follow prescribed system settings and thus do not have the mechanism to take actions based on the ever-changing process states. Intelligent AM, on the contrary, minimizes human inputs in the optimization process but relies on an automated process-level control mechanism to generate optimal design variables and adaptive system settings for improved end-product properties. In an ongoing INL Laboratory Directed Research and Development (LDRD) project, a novel artificial intelligence (AI)-based process control and optimization (PC\&O) methodology is being developed to intelligently control and optimize AM processes instead of the existing trial-and-error approach. This AI-based PC\&O methodology relies on Deep Reinforcement Learning (DRL) to build an online interaction mechanism in the process-informed design. However, one major challenge is that DRL requires a fast evaluation of the expected AM results. Therefore, the ROM can be employed by the DRL algorithm as it requires much less computational time than the full model.

There are two options to build the ROM with deep learning (DL) while preserving the ``physics'' of the physical model. The first one is physics-informed machine learning (PIML) \cite{karniadakis2021physics}, for example, using the physics-informed neural networks (PINN) \cite{raissi2019physics}. PINNs will solve the differential equations directly while enforcing the physics from the initial/boundary conditions and conservation equations in the loss function when training the deep neural networks. However, PINN works best for simple geometries and clearly-defined differential equations. In this application, the AM process will cause changing geometry due to the formation of a melt pool. Other challenges include modeling the moving laser power source and tracking the melt pool surface, or interface with the substrate. Therefore, it is difficult to implement the convective boundary condition. Moreover, PINN can only solve one instance of differential equations at once. In the MOOSE-based AM model, the thermal model uses a Gaussian point heat source that has multiple process variables to be controlled, such as laser power and laser scanning speed as described in Section \ref{section:methodologies}. Changing values of these process variables will result in a different set of equations that have to be re-trained by PINNs. These issues have motivated us to consider the second option, which is to build a data-driven ROM based on operator learning (OL) instead of solving the equations directly. OL learns an operator, which can solve a family of differential equations. In other words, the process variables in the Gaussian point heat source will be treated as inputs. When the differential equations change due to changes in this input, the OL-based ROM does not have to be re-trained. Moreover, an OL-based ROM is data-driven, making it easier to train because it does not solve the differential equations directly but uses the input-output paired training data from physical model simulations.

\section{Model Description}
\label{section:problem_definition}

The MOOSE-based AM model \cite{yushu2022directed} comprises two key components, namely the thermal model and the mechanical model. The former simulates the laser-induced heating of the powder material by solving the conservation of energy equation across the entire computational domain, while the latter solves the momentum conservation equation to compute the material stress. During the manufacturing process, heat conduction with a convective boundary is also considered and is represented by an equation that applies to the entire spatial domain $\Omega$. The heat conduction equation is written as follows:

\begin{align}    \label{eqn:heat-conduction}
        \rho c \left( T \right) \frac{\partial T }{\partial t} &= \nabla \cdot \kappa \left(T \right) \nabla T + Q \left( \bm{x}, t\right) \: \text{in} \: \Omega,\nonumber \\
        T &= \Bar{T} \: \text{on} \: \partial \Omega_{s, \text{bot}},\\
        - k \left( T \right) \nabla T \cdot \bm{n} &= h \left( T \right) \cdot \left(T - \Bar{T}_{\infty}\right) \: \text{on } \Gamma_{a, u} \nonumber
\end{align}
where $T$ is the temperature, $t$ is the time, $\bm{x} \in \mathbb{R}^{3}$ is the spatial location, $\bm{n} \in \mathbb{R}^{3}$ denotes the outward normal, $\rho$ is the material density, $c\left(T\right)$ is the temperature-dependent specific heat, $k \left( T \right) $ is the temperature-dependent thermal conductivity, and $h \left(T\right)$ is the temperature-dependent heat convection coefficient. At the bottom of the substrate ($\partial \Omega_{s, \text{bot}}$), the temperature is fixed at room temperature $\Bar{T}$. $\Bar{T}_{\infty}$ denotes the temperature far from the convective boundary, $\Gamma_{a, u}$ represents the interface of the growing deposited material. $Q \left( \bm{x}, t\right) $ is a time and spatially varying heat source that mimics the scanning laser beam used in the DED process. The heat source is a combination of the Gaussian point heat source model and the Gaussian line heat source model. The Gaussian heat source is defined as below:
\begin{equation}
  \hat{Q} (\bm{x},t)  =  \frac{2 \alpha \eta P}{\pi r^3} \exp \left(  - \frac{2 || \bm{x} - \bm{p}(t) ||^2}{r^2}  \right)
\end{equation}
where $P$ is the laser power, $\alpha$ is the equipment-related scaling factor, $\eta$ is the laser efficiency coefficient, $r$ is the effective radius of the laser beam, $v$ is the scanning velocity, and $\bm{p}(t) = vt$ is the time-varying scanning path. In this preliminary analysis, we assume the ROM takes five process variables as inputs, $P$, $\alpha$, $\eta$, $r$, and $v$. Table \ref{table:AM-process-variables} shows their nominal values and lower/upper bounds within which the ROM will be trained. Mechanical analysis is performed following thermal analysis in the material and the substrate. Due to the page limit, the mechanical model details will not be presented in this paper.

\begin{table}[htbp!]
\vspace{-2em}
	\centering
	\caption{Process parameters to be controlled in the AM process.}
	\vspace{1mm}
	\begin{tabular}{l c  c  c  c}
		\toprule
		Parameters & Symbols &  Nominal values & Bounds & Units\\
		\midrule
		Laser power  & $P$  & 300  & [250, 400] & W\\ 
		Scanning speed & $v$ & 0.01058 & [0.004, 0.020] & mm/ms\\
		Laser beam radius & $r$ & 0.3 &  [0.25, 0.40] & mm \\
		Laser efficiency coefficient & $\eta$ & 0.36  & [0.3, 0.4] & -\\
		Scaling factor  & $\alpha$ & 1.6 & [1.0, 2.0] & - \\
		\bottomrule
	\end{tabular}
	\label{table:AM-process-variables}
	\vspace{-1em}
\end{table}

MOOSE utilizes the finite element method (FEM) to solve differential equations. In each time step, the heat equation (\ref{eqn:heat-conduction}) is solved to obtain the temperature field for the entire space $\Omega$. To ensure the accuracy of the simulations, Adaptive Mesh Refinement is employed to dynamically adjust the mesh resolution of elements \cite{yushu2022directed}. This technique refines elements with a high prediction error and coarsens those with a low prediction error, resulting in an optimal mesh resolution for the simulation.

\section{METHODOLOGIES}
\label{section:methodologies}

\subsection{Deep Neural Networks}

In this work, a neural network (NN) is used as a supervised machine learning algorithm that learns to map labeled data between two sets. It comprises a multitude of computational units, referred to as \textit{neurons} or \textit{nodes}, organized into layers known as \textit{hidden layers} located between the input and output layers. A NN that features multiple hidden layers is referred to as a \textit{deep neural network (DNN)}. In a NN, neurons in the input layer are connected to those in the first hidden layer, with this connection extending to the output layer. Each of these connections is assigned a numeric value that indicates its importance, known as the \textit{weight}, denoted by $w$. If the output of one layer is used as input to the next layer, the network is referred to as a \textit{Feed-Forward Neural Network}.

A neural network utilizes an activation function to sequentially apply linear and non-linear transformations to input features to obtain output predictions. This function comprises two learnable parameters: weights $\bm{W}$ and hidden layer bias $\bm{b}$. However, the accuracy of the output prediction also relies on a set of user-defined parameters, commonly known as \textit{hyperparameters}. These parameters encompass the number of neurons per layer, the number of hidden layers, the learning rate, the optimization algorithm, and the type of activation function utilized. Appropriate selection of these hyperparameters can significantly influence the accuracy of the output prediction.

In a regular feed-forward architecture:
\begin{equation}    \label{eqn:FNN-output}
   \hat{y} \left(x\right) = \sigma \left(\cdots \sigma \left( \sigma \left(x \bm{W}_1 + \bm{b}_1\right) \bm{W}_{2} + \bm{b}_{2} \right)\cdots  \bm{W}_{L} + \bm{b}_{L} \right) 
\end{equation}

\begin{figure}[!htb]
	\centering
	\captionsetup{justification=centering}
	\includegraphics[width=0.7\textwidth]{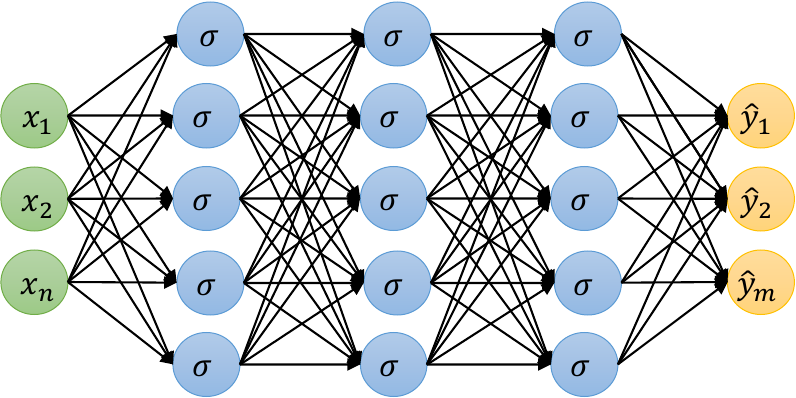}
	\caption[]{Illustration of a DNN with three hidden layers.}
	\label{fig:DNN-demo}
\end{figure}

The depth of a neural network is denoted by $L$, and it is determined by the number of hidden layers $(L-1)$ in addition to the output layer. $\sigma$ is the activation function. $\bm{W}_{l}$ is the matrix of weights for the $\bm{l}^{\text{th}}$ hidden layer. $\bm{b}_{l}$ is the vector of bias for the $\bm{l}^{\text{th}}$ hidden layer. $\hat{y} \left(x\right)$ is the prediction for input $x$. A typical DNN is depicted in Figure \ref{fig:DNN-demo}.

\subsection{Neural Operator}

In the field of scientific machine learning, the primary objective of employing DNNs is to learn mappings for finite-dimensional data, which in turn, allows them to perform regression or classification tasks. However, when utilizing these networks to solve partial differential equations (PDEs), the solution's spatial domain is typically discretized into a grid of specific resolution, with the training data generated using conventional numerical solvers. Unfortunately, the trained DNN may not generalize well to new data of different discretization schemes and resolutions, which necessitates the model's retraining with new structures and parameter optimization, resulting in a time-consuming and computationally expensive process. To address these issues, a new class of NNs has been proposed in \cite{kovachki2021neural} called \textit{Neural Operators}. These networks learn mappings between infinite-dimensional function spaces, allowing them to directly learn the operator instead of the solution instances of a PDE. Neural Operators can learn the operator directly, mapping the input function to the output function, making them discretization-invariant. This means that the discretization of the training data can be accomplished using various schemes and resolutions without affecting the prediction accuracy of the model. The training process is executed only once and the model can be used to predict any input data generated from different instances of the PDEs (e.g., by changing a coefficient, or the forcing term, etc.). In short, conventional data-driven ROMs try to learn the mapping from the model inputs to the model responses based on training data generated by the physical model, while Operator Learning (OL), such as a neural operator, tries to learn an operator of a differential equation.

The training data are arranged into input-output pairs that represent their function spaces. Let us denote a bounded open set as $D \in \mathbb{R}^{d}$ in which the function spaces $\mathcal{U}$ and $\mathcal{V}$ takes values in $\mathbb{R}^{d_{u}}$ and $\mathbb{R}^{d_{v}}$. Moreover, let $L^{\dagger}: \mathcal{V} \rightarrow \mathcal{U}$ be the non-linear map to be learned. Consider a dataset $\{ v_{i}, u_{i} \}_{j = 1}^{N}$ where $v_{i} \sim \mu$ samples are independent and identically distributed drawn from a probability distribution $\mu$. $u_{i} = L^{\dagger} \left( v_{i} \right)$ are points from output function. The goal is to build an approximation of $L^{\dagger} $ by constructing a parameter map.

\begin{equation}    \label{eqn:operator-approx}
    L_{\theta} : \mathcal{V} \rightarrow \mathcal{U}, \: \theta \in \Theta
\end{equation}

The parameters $\theta^{\dagger}$ from the finite-dimensional space $\Theta$ such that $L_{\theta} \approx L^{\dagger}$. We define a cost function $C: \mathcal{U} \times \mathcal{U} \rightarrow \mathbb{R}$ and use a minimizer to optimize it.

\begin{equation}
    \underset{\theta \in \Theta}{\text{min}} \: \mathbb{E}_{v \sim \mu} \left[ C\left( L\left(v, \theta\right),  L^{\dagger}\left(v\right)\right)\right]
\end{equation}

\begin{figure}[!htb]
	\centering
	\captionsetup{justification=centering}
	\includegraphics[width=0.7\textwidth]{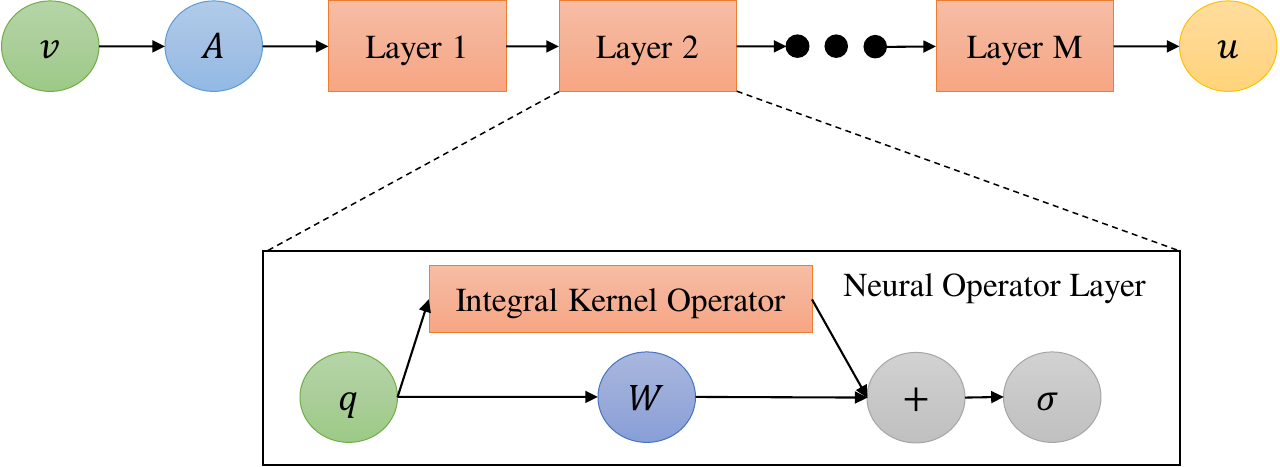}
	\caption[]{Neural operator illustration.}
	\label{fig:neural-operator-demo}
\end{figure}

The input function $v \left( x \right)$ is lifted to a higher dimensional representation $q \left( x \right)$ by a fully-connected neural network denoted as $A$ as shown in Figure \ref{fig:neural-operator-demo}. $q \left( x \right)$ is updated through the layers iteratively $q_{0} \rightarrow q_{1} \rightarrow q_{M} $ where $q_{i}$ for $i = 0, 1, \cdots, M -1$ values are taken from $\mathbb{R}^{d_{q}}$. $q_{0} \left( x \right) = A \left( v \left( x \right) \right)$ is first computed then each iteration is updated using a composition of a non-local integral operator $\mathcal{K}$ and a local, nonlinear activation function $\sigma$. Afterward, $q_{M} \left( x \right)$ is projected back to the output space by local transformation. Iterative updates are defined as follows:
\begin{equation}    \label{eqn:OL-iterative-updates}
    q_{m+1} \left( x \right) = \sigma \left( W q_{m} \left( x \right) +   \left( \mathcal{K} \left( v; \theta \right) q_{m} \right)  \left( x \right) \right), \: \forall x \in D
\end{equation}

Where $\mathcal{K}$ maps the linear operator $\mathcal{U}$ to a bounded space $\mathbb{R}^{d_{q}}$ parameterized by $\theta \in \Theta_{\mathcal{K}}$, $W: \mathbb{R}^{d_{q}} \rightarrow \mathbb{R}^{d_{q}}$ is a linear transformation, and $\sigma:\mathbb{R}\rightarrow \mathbb{R}$ is a non-linear activation function evaluated in a pointwise manner. The integral kernel operator $\mathcal{K}$ in equation (\ref{eqn:OL-iterative-updates}) is defined as follows:
\begin{equation}    \label{eqn:integral-kernel-operator}
     \left( \mathcal{K} \left( v; \theta \right) q_{m} \right)  \left( x \right) = \int_{D} \kappa \left( 
 x, y, q\left( x \right) ; \theta \right) q\left( y \right) \: dy, \: \forall x \in D
\end{equation}

Where $\kappa_{\theta}$ is a neural network parameterized by $\theta \in \Theta_{\mathcal{K}}$ and learned from data. Despite the fact the integral operator is linear, it can learn highly complex non-linear operators by using linear integral operators with non-linear activation functions. Neural operator layers can be graph neural operator, lower-rank neural operator, and Fourier neural operator (FNO) as demonstrated in \cite{kovachki2021neural}.

\subsection{Fourier Neural Operator}

If we drop the dependence on the input function $v \left( x \right)$ and require that $\kappa_{\theta} = \kappa_{\theta} \left( x - y \right)$, the integral operator becomes a convolution operator. The authors in \cite{li2020fourier} replaced the integral kernel operator in equation (\ref{eqn:integral-kernel-operator}) by convolution operator defined in the Fourier space and evaluated using Fast Fourier Transform (FFT), allowing for a fast and efficient computation approach. Let $\mathcal{F}$ be the Fourier transform of the a function $f: D \rightarrow \mathbb{R}^{d_{q}}$ and $\mathcal{F}^{-1}$ its inverse then:
\begin{equation}    \label{eqn:fourier-inverse-fourirer-def}
    \left( \mathcal{F} f \right)_{i} \left( t \right) = \int_{D} f_{i} \left( x \right) e^{- 2 j \pi \langle x, t \rangle} dx, \:  ( \mathcal{F}^{-1} f)_{i} \left( x \right) = \int_{D} f_{i} \left( t \right) e^{- 2 j \pi \langle x, t \rangle} dt
\end{equation}

For $i = 1, \cdots , d_{q}$. $j$ is the imaginary number. Replace $\kappa \left( 
 x, y, q\left( x \right) ; \theta \right) q\left( y \right)$ by $\kappa_{\theta} \left( x - y \right)$ in equation (\ref{eqn:integral-kernel-operator}) and apply convolution theorem, it results:
 \begin{equation}    \label{eqn:covolution-Fourier}
     \left( \mathcal{K} \left( v ; \theta \right) q_{m} \right) \left( x \right) = \mathcal{F}^{-1} \left( \mathcal{F} (\kappa_{\theta}) \cdot \mathcal{F} \left(q_{m} \right) ) \right) \left( x \right) , \: \forall x \in D
 \end{equation}

 Now we can define the Fourier integral operator as follows:
 \begin{equation}    \label{eqn:Fourier-integral-operator}
     \left( \mathcal{K} \left(\theta \right) q_{m} \right) \left( x \right) = \mathcal{F}^{-1} \left( R_{\theta} \cdot \mathcal{F}\left(q_{m} \right) ) \right) \left( x \right), \forall x \in D 
 \end{equation}

Where $R_{\theta}$ is the Fourier transform of the function $\kappa$. Let $\kappa$ be periodic, the Fourier series expansion will produce discrete frequency modes $t \in \mathbb{Z}^{d}$, then the  Fourier series is truncated at the maximum number of modes $t_{\text{max}}$ representing  $R_{\theta}$. In each layer shown in Figure \ref{fig:neural-operator-demo}, a Fourier transform is implemented for the transformed input $q$. Next, the higher-order modes are filtered out before an inverse Fourier transform is performed. The output of the last Fourier layer is projected back to the target dimension by a neural network. To make a prediction, the trained FNO model is used to compute the Fourier coefficients of the desired quantity of interest (QoI). Those coefficients are converted back to the original data domain using the inverse Fourier transform.

\section{RESULTS}
\label{section:results}

\subsection{ROMs Results}

\begin{figure}[!htb]
	\centering
	\captionsetup{justification=centering}
	\includegraphics[width=1.0\textwidth]{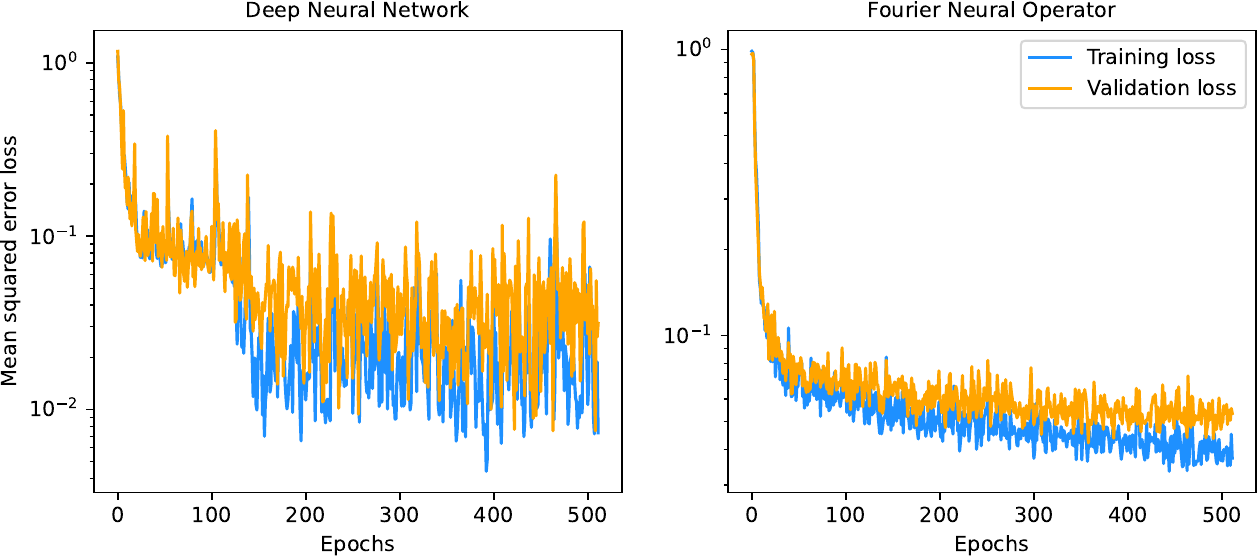}
	\caption[]{Training and validation MSE loss functions for the DNN and FNO models.}
	\label{fig:FNO-DNN-learning-curve}
\end{figure}

Both conventional DNN and FNO were employed as data-driven ROMs to the high-fidelity MOOSE-based AM model, to be used as a fast-running and accurate ROM. Note that here ``conventional DNN'' means that OL is not used. FNO also uses DNNs, but in a different way because the OL technique is employed. The training inputs include five process variables listed in Table \ref{table:AM-process-variables}, while the bead volume and maximum melt pool temperature are treated as outputs, or QoIs. The MOOSE model was run to generate a total of 500 samples, but some simulations were removed because they didn't result in a melt pool. As a result, the final sample size was reduced to 470. 80\% of the dataset was utilized for training and 10\% for both testing and validation. The Fourier integral operator was created by stacking 4 Fourier layers with 5 modes ($t_{\text{max}} = 5$) each with the GeLU activation function. The models were trained using the Adam optimizer for 512 epochs with a learning rate of $0.001$ for FNO and $0.007$ for DNN. The DNN architecture consists of 5 hidden layers with $(150,300,500,300,150)$ neurons in each layer and the ReLU activation function. The hyperparameters of both ROMs were optimized through grid search, with the learning curve shown in Figure \ref{fig:FNO-DNN-learning-curve} displaying the evolution of the mean squared error (MSE) loss function with epochs. We can notice that both validation and training losses for both ROMs are close, eliminating the possibility of over-fitting. DNN loss function oscillates higher than FNO due to the higher learning rate used.

\begin{figure}[!htb]
	\centering
	\captionsetup{justification=centering}
	\includegraphics[width=1.0\textwidth]{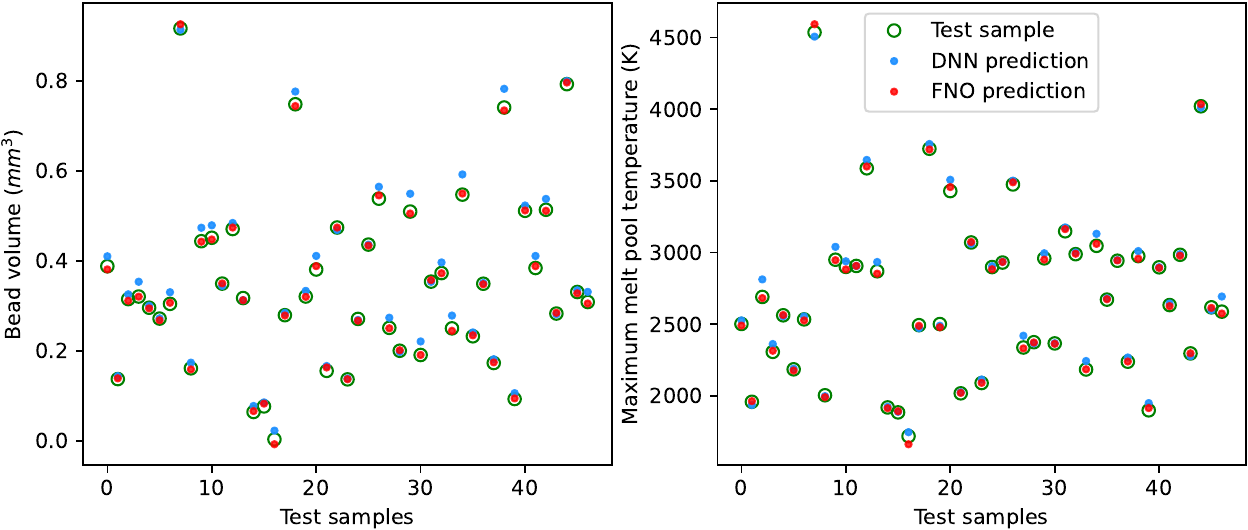}
	\caption[]{ROMs prediction compared with the test data.}
	\label{fig:FNO-DNN-pred-comparison}
\end{figure}

\begin{figure}[!htb]
	\centering
	\captionsetup{justification=centering}
	\includegraphics[width=1.0\textwidth]{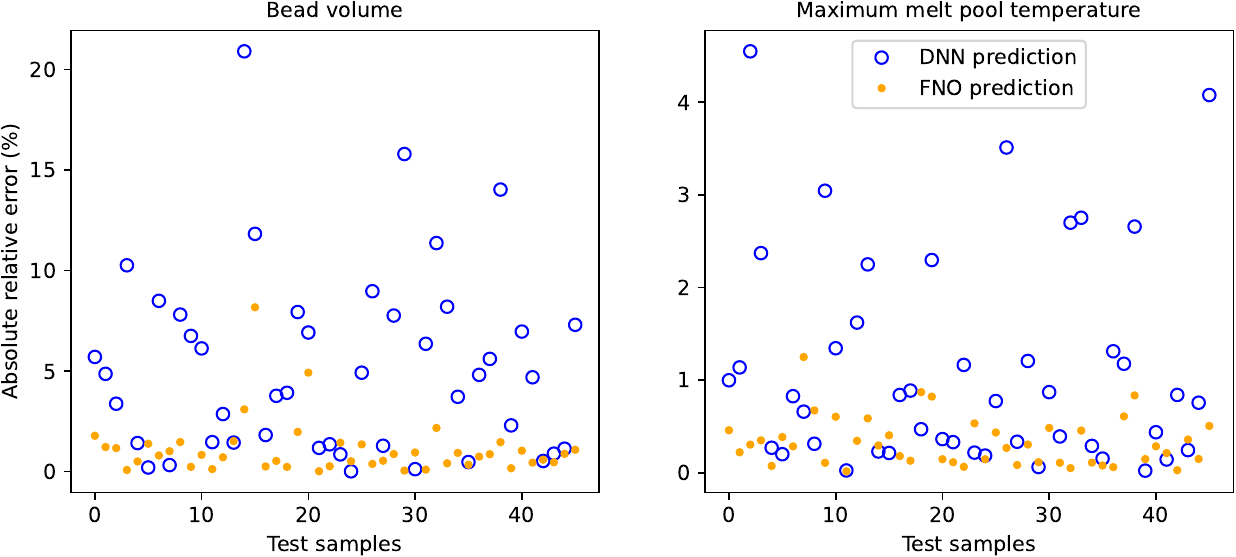}
	\caption[]{Absolute relative error of the ROMs prediction with respect to the test data.}
	\label{fig:FNO-DNN-absolute-error}
\end{figure}

Figure \ref{fig:FNO-DNN-pred-comparison} illustrates the predicted bead volume and maximum melt pool temperature, compared with the testing data. It is evident that the FNO model's predictions for all 47 test samples agree well with the true data points, indicating high generalizability and precise forecasts. The FNO model's training time was $276.85$ seconds, and it can make instantaneous predictions for any new instance of the input parameters. In contrast, the DNN has a faster training time of $69.32$ seconds, and its predictions also exhibit good agreement with true values. However, when compared to FNO, DNN's predictions are less accurate. This is demonstrated in Figure \ref{fig:FNO-DNN-absolute-error}, where the absolute relative errors of DNN for the majority of the test samples of both bead volume and maximum melt pool temperature are larger than those of the FNO model. Additionally, Table \ref{table:FNO-DNN-performance_comparison} shows that the root mean squared errors (RMSE) of both predicted QoIs for the FNO model are smaller than those of the conventional DNN. Furthermore, Table \ref{table:FNO-DNN-performance_comparison} displays the coefficient of determination ($R^{2}$), which represents how well the surrogate model approximates the true data. The $R^{2}$ value for the FNO model is closer to one than that of the DNN, indicating that the neural operator better fits the true model than the DNN.

\begin{table}[htbp!]
        \centering
        \caption{Performance comparison of the conventional DNN and FNO-based ROMs.}
        \begin{tabular}{c c c c c }
        \hline
         \multirow{1}{*}{QoI}  & \multicolumn{2}{c}{FNO} & \multicolumn{2}{c}{DNN} \\
        \cline{2-5}  & RMSE & $R^{2}$ & RMSE & $R^{2}$ \\
        \hline
        Bead volume & 0.004058    &  0.99954 &  0.019712  & 0.98921 \\
        \hline
        Maximum melt pool temperature & 15.497 & 0.99927 & 42.327  & 0.99458 \\
        \hline
        \end{tabular}
        \label{table:FNO-DNN-performance_comparison}
\end{table}

\subsection{FNO Time Series Prediction Results}

\begin{figure}[!htb]
	\centering
	\captionsetup{justification=centering}
	\includegraphics[width=0.9\textwidth]{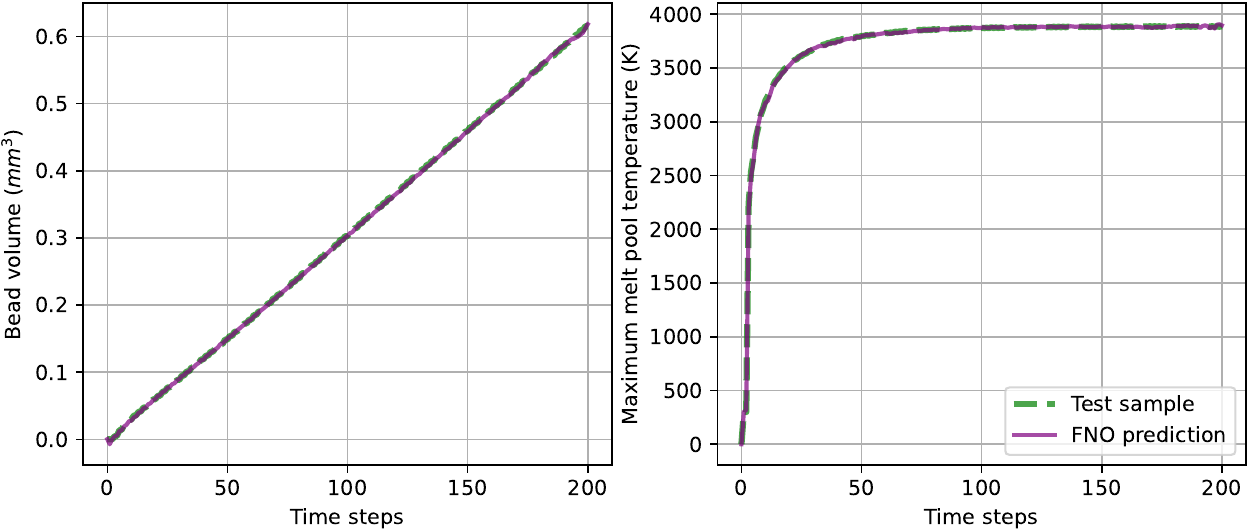}
	\caption[]{An example of the comparison between FNO prediction and AM model simulations for the time-dependent bead volume and maximum melt pool temperature.}
	\label{fig:FNO-time-series-pred}
\end{figure}

\begin{figure}[!htb]
	\centering
	\captionsetup{justification=centering}
	\includegraphics[width=0.9\textwidth]{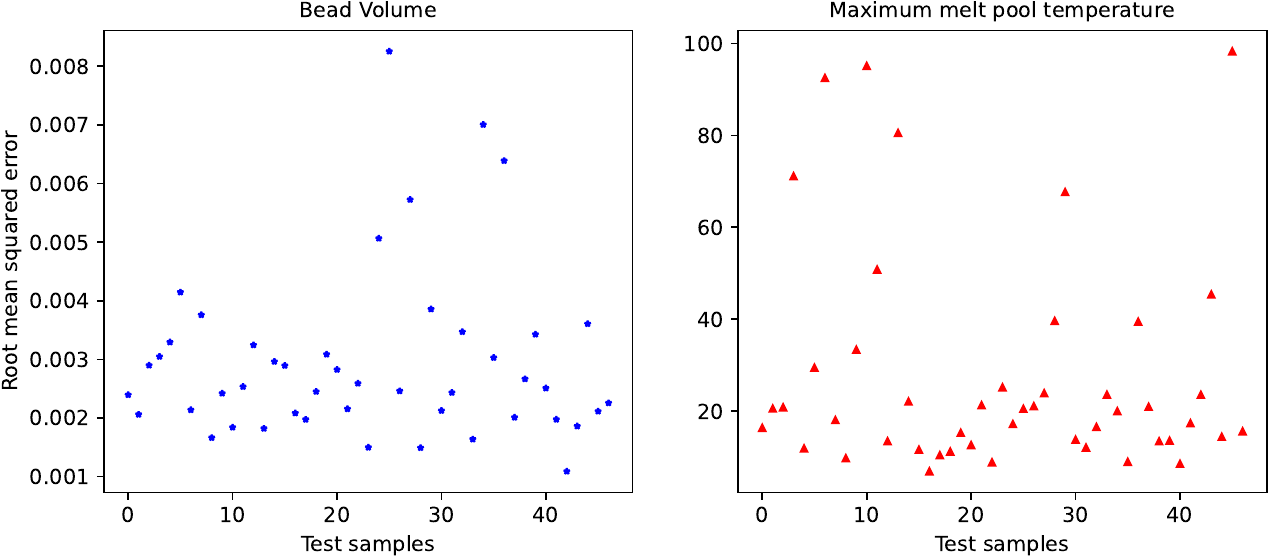}
	\caption[]{RMSE of the FNO time series prediction.}
	\label{fig:FNO-time-series-pred-rmse}
\end{figure}

Even though the MOOSE-based AM model can produce time-dependent simulations, the conventional DNN-based ROM model is limited to predicting scalar QoIs, which are bead volume (generally at the end of simulation) and maximum melt pool temperature. In contrast, FNO is capable of learning the operator itself and can be trained to forecast time series outputs. The input layer contains constant parameter values from Table \ref{table:AM-process-variables} in each sample, while the output layer contains bead volume and melt pool temperature values over the whole time series, in this case, at 200 time steps. To accurately capture the details of the time series QoIs, we used 50 modes in the FNO model as time-dependent outputs are more challenging to fit than scalar outputs. Figure \ref{fig:FNO-time-series-pred} displays the FNO time-series prediction for one test sample, which closely aligns with the true data from the MOOSE model. The error values for both QoIs in the test dataset are small, as shown in Figure \ref{fig:FNO-time-series-pred-rmse}, indicating that the FNO predictions for other test samples are reliable and accurate. The larger RMSE value for temperature than volume is attributed to the difference in scale between temperature and volume values.

\section{CONCLUSIONS}
\label{section:conclusions}

In this study, we employed operator learning (OL) as a reduced order modeling (ROM) approach for the MOOSE-based advanced manufacturing (AM) model to predict both the bead volume and maximum melt pool temperature. To build a fast-running and accurate ROM for the expensive high-fidelity physics-based AM model, we utilized the Fourier neural operator (FNO) that can learn a family of equations, and we compared its results with those of a conventional deep neural network (DNN)-based ROM. Our analysis demonstrated that the FNO model is an accurate and efficient ROM, which can effectively substitute the expensive AM model for multi-query tasks that involve repeated runs of the computational model. For instance, it can be utilized to optimize the control process in AM using deep reinforcement learning. While we observed that the conventional DNN model was faster to train than FNO due to its much simpler structure, it was less accurate and could not generalize as effectively as FNO for new datasets. We also trained FNO to forecast time series temperature and volume and obtained accurate predictions for time-dependent data. In general, operator learning methods outperform DNN in simulating complex processes such as AM, and they can predict both single values and time series of the quantities of interest.



\section*{ACKNOWLEDGEMENTS}

This work was supported through the INL Laboratory Directed Research \& Development (LDRD) Program under DOE Idaho Operations Office contract no. DE-AC07-05ID14517. This research made use of Idaho National Laboratory computing resources which are supported by the Office of Nuclear Energy of the U.S. Department of Energy and the Nuclear Science User Facilities under Contract No. DE-AC07-05ID14517.

\setlength{\baselineskip}{12pt}
\bibliographystyle{mc2023}
\bibliography{mc2023.bib}

\end{document}